\documentclass[10pt,conference]{IEEEtran} 
\IEEEoverridecommandlockouts
\usepackage{cite}
\usepackage{amsmath,amssymb,amsfonts}
\usepackage{algorithmic}
\usepackage{graphicx}
\usepackage{textcomp}
\usepackage{xcolor}
\usepackage{booktabs}
\usepackage{multirow}
\usepackage{subcaption}
\usepackage[linesnumbered,ruled,vlined]{algorithm2e}

\def\BibTeX{{\rm B\kern-.05em{\sc i\kern-.025em b}\kern-.08em
    T\kern-.1667em\lower.7ex\hbox{E}\kern-.125emX}}

\begin{document}
\setlength{\abovecaptionskip}{0.2cm}
\setlength{\belowcaptionskip}{-0.2cm}
\setlength{\abovedisplayskip}{3pt}
\setlength{\belowdisplayskip}{3pt}



\title{A Spatiotemporal Stealthy Backdoor Attack against Cooperative Multi-Agent Deep Reinforcement Learning\thanks{Accepted by IEEE Globecom 2024.}}


\author{
\IEEEauthorblockN{Yinbo Yu\IEEEauthorrefmark{1}\IEEEauthorrefmark{2},
Saihao Yan\IEEEauthorrefmark{1},
Jiajia Liu\IEEEauthorrefmark{1}}
\IEEEauthorblockA{\IEEEauthorrefmark{1}School of Cybersecurity, Northwestern Polytechnical University, Xi'an 710072, China}
\IEEEauthorblockA{\IEEEauthorrefmark{2}Research \& Development Institute of Northwestern Polytechnical University in Shenzhen, Shenzhen 518057, China}

\IEEEauthorblockA{Email: \{yinboyu, liujiajia\}@nwpu.edu.cn, yansaihao@mail.nwpu.edu.cn}
\\[-4.5ex]
}

\maketitle

\begin{abstract}
Recent studies have shown that cooperative multi-agent deep reinforcement learning (c-MADRL) is under the threat of backdoor attacks.
Once a backdoor trigger is observed, it will perform abnormal actions leading to failures or malicious goals.
However, existing proposed backdoors suffer from several issues, e.g., fixed visual trigger patterns lack stealthiness, the backdoor is trained or activated by an additional network, or all agents are backdoored. To this end, in this paper, we propose a novel backdoor attack against c-MADRL, which attacks the entire multi-agent team by embedding the backdoor only in a single agent. Firstly, we introduce adversary spatiotemporal behavior patterns as the backdoor trigger rather than manual-injected fixed visual patterns or instant status
and control the attack duration. This method can guarantee the stealthiness and practicality of injected backdoors. Secondly, we hack the original reward function of the backdoored agent via reward reverse and unilateral guidance during training to ensure its adverse influence on the entire team. We evaluate our backdoor attacks on two classic c-MADRL algorithms VDN and QMIX, in a popular c-MADRL environment SMAC. The experimental results demonstrate that our backdoor attacks are able to reach a high attack success rate (91.6\%) while maintaining a low clean performance variance rate (3.7\%).
\end{abstract}


\section{Introduction}
Effectively combining the perception and decision-making ability of deep reinforcement learning (DRL) with the distributed characteristics of multi-agent systems, cooperative multi-agent deep reinforcement learning (c-MADRL) has been applied to cooperative games \cite{samvelyan2019starcraft},  communication \cite{10437779}, and other fields \cite{oroojlooy2023review}. However, recent studies have shown that DRL is facing the security threat of backdoor attacks \cite{9218663, yang2019design, ijcai2021p509, ashcraft2021poisoning}. Attackers usually implant the backdoors by poisoning the observations, actions, or rewards of a victim DRL agent during training. A backdoored model is characterized by outputting normal behaviors under normal inputs, but malicious behaviors expected by the attacker when a specified backdoor trigger appears in the inputs. Therefore, c-MADRL, as a branch of DRL, is also under the threat of backdoor attacks.

Compared with backdoor attacks against DRL, conducting backdoor attacks against c-MADRL receives some extra challenges.
Firstly, it is easy to implant backdoor attacks in all agents, but this attack is expensive to inject and less stealthy. Hence, it is important to poison as few agents as possible. However, the attack effects generated from a few agents may be easily overwhelmed due to mutual dependency \cite{oroojlooy2023review} among c-MADRL agents.
So far, there are only a few studies \cite{10437779, MARNet, chen2022backdoor, zheng2023one4all} on backdoor attacks against c-MADRL. While attacks \cite{MARNet, 10437779} directly poison all agents, attacks \cite{chen2022backdoor, zheng2023one4all} aim to poison one agent to attack the entire system, but ignore agent mutual dependency, limiting the attack efficiency.
Secondly, most existing attacks use a specific instant trigger, e.g., spectrum signals \cite{10437779}, visual patterns \cite{MARNet}, distance \cite{zheng2023one4all}, and trigger policy \cite{chen2022backdoor}, which can be detected by anomaly classifiers \cite{yu2023spatiotemporal}, thus having poor concealment. Hence, hiding triggers in a series of timing states is more suitable for c-MADRL's sequential decision-making and can achieve better concealment. While existing backdoor attacks are useful for c-MADRL, they have not addressed these challenges well.

To solve the above challenges, in this paper, we propose a novel stealthy backdoor attack against c-MADRL, which can disrupt the entire multi-agent team by implanting the backdoor in a single agent without modifying its original network architecture. Firstly, most of the existing c-MADRL algorithms, such as VDN \cite{VDN}, QMIX \cite{QMIX} and COMA \cite{COMA}, use a recurrent neural network (RNN) to remember the past information and combine it with the current observation to make effective decisions, so as to overcome the partial observability \cite{DRQN}. However, this allows attackers to hide the backdoor triggers in unobservable states \cite{yu2022temporal, yu2023spatiotemporal}. Therefore, unlike existing methods that use instant triggers \cite{9218663}, \cite{yang2019design}, \cite{MARNet}, \cite{chen2022backdoor}, we carry out our backdoor attacks using a spatiotemporal behavior pattern trigger hidden in a series of observations and a controllable attack period. So, the attacker can act as a moving object (e.g., an enemy unit in Starcraft) to perform a series of specific actions around the backdoored agent in a short time to control the spatial and temporal state dependencies with the agent, thereby activating the backdoor. This trigger has proven to be highly stealthy and difficult to detect with existing detection methods in our previous studies \cite{yu2022temporal, yu2023spatiotemporal}.

Secondly, in c-MADRL, the behaviors of an agent are influenced not only by the running environment but also by the behaviors of other agents.
Therefore, an agent taking a different action may influence other agents' decisions in the next few steps. Based on this feature, we design a reward hacking method for injecting a backdoor into a c-MADRL agent.
The method includes two reward items: the former encourages the backdoored agent to perform the actions that have a long-term malicious impact on the entire team; the latter uses a unilateral influence filter \cite{li2023attacking} to guide the backdoored agent to quickly find the actions that can mislead its teammates to producing non-optimal actions. Given a pre-trained clean c-MADRL model, we use the above spatiotemporal pattern trigger and the reward hacking method to retrain the model and deploy it in the multi-agent team.

Our backdoor attack can be applied to team competition systems (e.g., MOBA games) or collaborative control systems (e.g., connected vehicles autonomous driving), where the attacker acts as a normal object in the observation of c-MADRL agents and performs specific actions to activate the backdoor to lead the c-MADRL team to have a low performance or fail.
We evaluate our backdoor attacks against two commonly used c-MADRL algorithms (VDN and QMIX) in SMAC \cite{samvelyan2019starcraft}, a popular c-MADRL environment. The experimental results demonstrate that our backdoor attacks are able to reach a high attack success rate (91.6\%) while maintaining a low clean performance variance rate (3.7\%).

\section{Related Work}
Most of the existing studies on backdoor attacks focuse on deep neural networks (DNN) \cite{DBLP:journals/tnn/LiJLX24}.
Due to the representation ability and feature extraction ability inherited from DNN, DRL also inevitably faces this threat.
Kiourti \textit{et al.} \cite{9218663} propose a backdoor attack method against DRL with image patch triggers. During training, when the model's input contains a trigger and the output is the target action or a random one, they maximize the corresponding reward value to achieve a targeted or untargeted attack.
Yang \textit{et al.} \cite{yang2019design} investigate the persistent impact of backdoor attacks against LSTM-based PPO algorithm \cite{schulman2017proximal}. When an image patch trigger appears at a certain time step, the backdoored agent persistently goes for the attacker-specified goal instead of the original goal.
Ashcraft \textit{et al.} \cite{ashcraft2021poisoning} use in-distribution triggers and multitask learning to train a backdoored agent.
Wang \textit{et al.} \cite{ijcai2021p509} study backdoor attacks against DRL in two-player competitive games. The opponent's actions are used as a trigger to switch the backdoored agent to fail fast.
Our previous works \cite{yu2022temporal, yu2023spatiotemporal} studied a new temporal and spatial-temporal backdoor trigger to DRL, which has shown high concealment.

So far, only a few studies have been done on backdoors in c-MADRL. Chen \textit{et al.} \cite{MARNet} first studied backdoor attacks against c-MADRL. They consider both out-of-distribution and in-distribution triggers, and use a pre-trained expert model to guide the selection of actions of poisoned agents. Besides, they modify the team reward to encourage poisoned agents to exhibit their worst actions. Their method implants backdoors in all agents, and every agent that observes the trigger will perform instant malicious action.
Chen \textit{et al.} \cite{chen2022backdoor} propose backdoor attacks that affect the entire multi-agent team by targeting just one backdoored agent. During training, they introduce a random network distillation module and a trigger policy network to guide the backdoored agent on what actions to select and when to trigger the backdoor, thereby enabling the agent to adversely affect other normal agents. During execution, the backdoored agent needs to rely on the trigger policy network to determine whether to trigger the backdoor.
Zheng \textit{et al.} \cite{zheng2023one4all} also implanted a backdoor into one of the agents in a multi-agent team, and use distance as a condition to trigger the backdoor. When the trigger condition is met, the backdoored agent performs a specified action which leads to the failure of the team task.
These existing backdoor attacks are effective against c-MADRL, but they either poison all agents, require additional networks to activate backdoors, or do not take into account the influence between agents.

\section{Threat Model}
\subsection{Problem Definition}
In this section, we logically represent our backdoor attacks against c-MADRL as a decentralized partially observable Markov decision process (Dec-POMDP) which consists of a tuple $\langle \{N, S, O, A, T, \{R^b, R^c\}, \gamma\rangle$.
\begin{itemize}
    \item $N:=\{1, ..., n\}$ represents the set of team agents in c-MADRL. We specify agent $k$ to implant the backdoor, and the other agents are clean agents without backdoors.
    \item $S$ represents the global environmental state space. Following the centralized training and decentralized execution (CTDE) paradigm \cite{VDN}\cite{QMIX}, $s_{t}\in S$ is only employed during training and not during testing (execution).
    \item $O:=O_1\times ... \times O_{n}$ represents the local observations of all agents. The individual observation $o_{i, t} \in O_i$ of each agent $i$ at time step $t$ serves as an input for the policy network of the agent.
    \item $A:=A_1\times ... \times A_{n}$ represents the joint actions of all agents. All clean agents and the backdoored agent use $\pi^c(a_{i, t}|o_{i, t}):O_i\rightarrow A_i$ and $\pi^b(a_{k, t}|o_{k, t}):O_i\rightarrow A_i$  to select action, respectively,  where $a_{i,t} \in A_i$ denotes the selected action of each individual agent.
    \item $T: S\times A\rightarrow S$ represents environmental state transition function. Given state $s_t\in S$ and joint action $\textbf{a}_t\in A$ at time step $t$, $T(s_{t+1}|s_t, \textbf{a}_t)$ computes the probability of transitioning to state $s_{t+1}\in S$ at time step $t+1$. Besides, depending on $T$, we can use $F(o_{i, t+1}|o_{i, t}, \textbf{a}_t)$ to represent the observation transition of agent $i$.
    \item $R^b, R^c: S\times A\times S\rightarrow \mathbb{R}$ represents the reward function for both the backdoored agent and clean agents after executing a joint action $\textbf{a}_t\in A$ that transitions the state from $s_t\in S$ to $s_{t+1}\in S$. The design of $r^b_t\in R^b$ plays a significant role in the success of backdoor attacks.
    \item $\gamma$ is the temporal discount factor where $0\leq \gamma < 1$.
\end{itemize}

We only inject the backdoor in a single agent to ensure the stealthiness and practicality of backdoor attacks. The backdoored agent behaves normally like a clean agent in the absence of an attacker-specified trigger within its observation. However, once the trigger appears, it selects disruptive behavior that influences the other clean agents and ultimately leads to the failure of the team.

\subsection{Attacker’s Capacities and Goals}
\textbf{Attacker’s Capacities}. For backdoor attacks against c-MADRL, we consider two parties (i.e., multiple users and a single attacker) and two possible attack scenarios. The first scenario is that the attacker participates in the multi-agent distributed system as one of the users. Based on the trained clean c-MADRL model, she further retrains it as a backdoored model and deploys it in the team as a hidden ``traitor''. In the second scenario, a user outsources the training of agent models to a third-party platform due to a lack of training skills, simulation environments, or computing resources. The attacker, acting as the contractor for the third-party platform, injects the backdoor into the user's model. In both scenarios, the attacker has the ability to modify some training data, including observations and corresponding rewards, but not change the model's network structure.

\textbf{Attacker’s Goals}. With the above capabilities and limitations, the attacker's main goal is to attack the entire multi-agent team by triggering the backdoor in a single agent during execution. Moreover, the implanted backdoor needs to be effective and stealthy. Specifically, if the backdoor trigger is present, the backdoored agent is capable of behaving maliciously or abnormally to disrupt the team, otherwise it is capable of behaving normally like a clean agent. Besides, the backdoor triggers should be concealed, occur as infrequently as possible, and have a small poisoning rate.

\section{The Proposed Backdoor Attack Method}

\begin{figure}[t]
    \captionsetup{font=small}
    \centering
    \includegraphics[width=\linewidth]{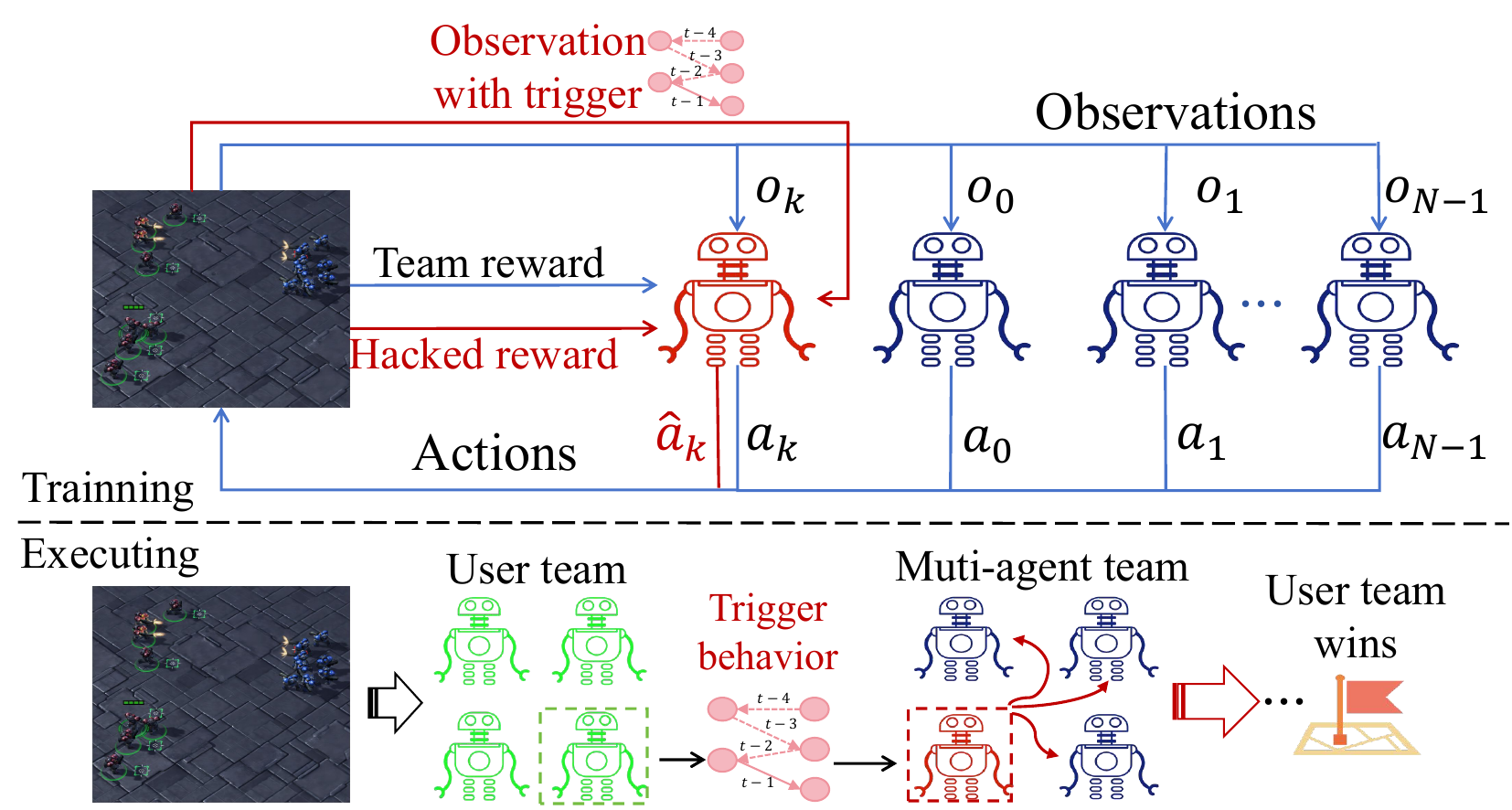}
    \caption{The framework of the proposed backdoor attack.}
    \label{fig:overview}
    \vspace{-10pt} 
\end{figure}

In this section, we formulate our designed spatiotemporal backdoor trigger and the reward hacking method based on reward reversing and unilateral influence and describe the training procedure of the backdoored model. Fig. \ref{fig:overview} shows the framework of the proposed backdoor attack.

\subsection{Spatiotemporal Backdoor Trigger}
In many input-driven multi-agent systems, such as MOBA games \cite{samvelyan2019starcraft} and autonomous driving \cite{yu2023spatiotemporal}, the observations of each agent typically include four types of information: its own state, teammates' state, internal environmental information, and information from external inputs. The information from external inputs is not controlled by the environment or the agents' decisions and may be exploited by malicious attackers as backdoor triggers. For example, in a MOBA game (e.g., Starcraft), an attacker can control an enemy unit to perform a series of special actions in a short time to activate the backdoor attack.
In this section, we present a spatiotemporal behavior pattern trigger that represents a set of specific spatial dependencies between the attacker's unit and the backdoored agent, as well as a set of specific temporal behaviors of the attacker's unit. We use a logical formula to represent the \textit{spatial dependencies} and a set of the enemy unit's controllable actions to represent the \textit{temporal behaviors}. First, given a position state $s^b$ of the unit controlled by the backdoored agent and $s^e$ of the attacker's unit, we use $\psi := g(s^b, s^e) \thicksim C$ to describe the spatial constraint between them at a given time step, where $g \in \{+, -, \times, \div\}$ is an operator, $\thicksim \in \{ >, \geq, <, \leq, \equiv, \neq\}$ is a relator, and $C \in \mathbb{R}$ is a constant. Specifically, we define the trigger as follows:

\textbf{Definition 1.} Given consecutive position states over a short period length $\mathbb{N}^t$ ending at time step $t$, a \textbf{spatiotemporal behavior trigger} $\mathcal{T} := (\Psi, \zeta)$ can be defined using a logic formula $\Psi$ representing the spatial constraints $\Omega$ of these states and a set $\zeta$ of controllable actions:
\begin{equation}
\Psi:=\psi_{(i,p)}|\psi_{(i,p)}\otimes\psi_{(j,q)}|ite(\psi_{(i,p)},\psi_{(j,q)},\psi_{(k,s)})
\end{equation}
\begin{equation}
    \zeta:=(a^e_{t-\mathbb{N}^t+1},a^e_{t-\mathbb{N}^t+2},\cdots,a^e_t)
\end{equation}

\noindent  where $\psi_{(i,p)}, \psi_{(j,q)}, \psi_{(k,s)} \in \Omega$, $i, j, k \in \mathbb{N}$ are time steps, and $p, q, s $ are the position features of attacker's unit relative to backdoored agent; $\otimes \in \{ \vee, \wedge\}$ is a Boolean operator (\textit{i.e.}, ``or'' or ``and'');  $ite$ represents $\Psi$-assignment, e.g., $\Psi := ite(\psi_1, \psi_2, \psi_3)$ means if $\psi_1$ is true, $\Psi := \psi_2$; otherwise $\Psi := \psi_3$; $a^e_t$ is an action executed by the attacker’s unit.

\textbf{Example 1.} Take Starcraft for example, given a trigger $\mathcal{T} := (\Psi, \zeta)$, where $\Psi(\mathbb{N}^t=5) := \psi_{(t-4,x)} \wedge \psi_{(t-4,y)} \wedge \psi_{(t-3,x)} \wedge \psi_{(t-3,y)} \wedge \psi_{(t-2,x)} \wedge \psi_{(t-2,y)} \wedge \psi_{(t-1,x)} \wedge \psi_{(t-1,y)} \wedge \psi_{(t,x)} \wedge \psi_{(t,y)}$, and $\zeta := (a^e_{t-4}, a^e_{t-3}, a^e_{t-2}, a^e_{t-1}, a^e_{t})$, where $\psi_{(t-4,x)} = \psi_{(t-2,x)} = \psi_{(t,x)} := 8.8 < x^e - x^b < 9.0$, $\psi_{(t-3,x)} = \psi_{(t-1,x)} := 6.1 < x^e - x^b < 6.3$, $\psi_{(t-4,y)} = \psi_{(t-3,y)} = \psi_{(t-2,y)} = \psi_{(t-1,y)} = \psi_{(t,y)}:= -0.1 < y^e - y^b < 0.1$; $a^e_{t-4}$ represent the attacker's enemy unit moving westward, $a^e_{t-3}$ is moving eastward, $a^e_{t-2}$ is moving westward, and $a^e_{t-1}$ is moving eastward.

With this backdoor (shown in Fig. \ref{fig:overview}), a user in the user team can perform its unit following the trigger to activate the backdoor in the backdoored agent of the multi-agent team. This agent will generate malicious actions to lead the multi-agent team to quickly lose the game.

\subsection{Reward Hacking}
Firstly, to guarantee the clean performance of the backdoored agent in the trigger-free environment, at each time step $t$, we use the original team reward $R_t$ in c-MADRL to train the backdoored model, where $R_t \in [R_{min}, R_{max}]$. During the attack period after the trigger completely appears, We reverse the original environmental reward $R_t$ of the backdoored agent to the following form:
\begin{equation}
    r_t^e = R_{max} + R_{min} - R_t
\end{equation}

The purpose of $r_t^e$ is to stimulate bad actions and punish good actions. At the same time, $r_t^e$ retains the same value range as $R_t$, which can reduce the damage to the clean performance of the backdoored agent.

In the c-MADRL system, all agents typically interact with each other. Due to this mutual influence, the sphere of influence of $r_t^e$ may be limited only to the backdoored agent itself rather than the entire system during the attack period. To address this problem, we further design a reward term based on the unilateral influence filter \cite{li2023attacking} which can eliminate the detrimental influence from other agents to the backdoored agent and only enable the influence from the backdoored agent to other agents as follows:

\vspace{-2mm}
\begin{footnotesize}
\begin{equation}
    \begin{aligned}r_{t}^{I} &= \sum_{ i=1, i\neq k}^{n}d\left(\hat{a}_{i, t+1},a_{i, t+1}\right) \\ &= \sum_{i=1, i\neq k}^{n}d\big(\pi^c\big(\hat{o}_{i, t+1}\big),\pi^c\big(o_{i, t+1}\big)\big) \\ &= \sum_{i=1, i\neq k}^{n}d\Big(\pi^c\Big(F(o_{i, t},\hat{a}_{k, t},\boldsymbol{a}_{-k, t})\Big),\pi^c\Big(F(o_{i, t},a_{k, t},\boldsymbol{a}_{-k, t})\Big)\Big) \end{aligned}
\label{equ:un}
\end{equation}
\end{footnotesize}

\vspace{-2mm}

\noindent where ${o}_{i, t+1}$ and ${a}_{i, t+1}$ denote the observation and action of the clean agent $i$ at the time step $t+1$, respectively, when the backdoored agent $k$ takes a normal action $a_{k, t}$ at $t$; $\hat{o}_{i, t+1}$ and $\hat{a}_{i, t+1}$ denote the different observation and action taken by the agent $i$ if the agent $k$ takes a malicious action $\hat{a}_{k, t}$ at $t$; $\boldsymbol{a}_{-k, t}$ represents the joint actions taken by all clean agents at time step $t$; $F(\cdot)$ represents the observation transition function which depends on  $T$; $d(\cdot , \cdot)$ is a distance metric function with $d\left(\hat{a}_{i, t+1},a_{i, t+1}\right) = 0$ if $\hat{a}_{i, t+1} = a_{i, t+1}$ and $d\left(\hat{a}_{i, t+1},a_{i, t+1}\right) = 1$ otherwise. During our backdoor injection process, the policy $\pi^c$ of the clean agents is frozen. Hence, we can only induce the change of their actions $a_{i, t+1}$ by changing $o_{i, t+1}$. Using $F(\cdot)$, the backdoored agent can select action $\hat{a}_{k,t}\thicksim \pi^b(o_{k,t})$ different from the action $a_{k, t}\thicksim \pi^c(o_{k, t})$ to change the observation $o_{i, t+1}$ into $\hat{o}_{i, t+1}$.

To calculate $r_{t}^{I}$, if the current time step $t$ is in the attack period, we record and save the global state $s_t$ and the local observation $o_{i, t}$ of all clean agents.
First, we let the backdoored agent $k$ select action $a_{k, t} \thicksim \pi^c(o_{k, t})$ and each clean agent $i$ select action $a_{i, t} \thicksim \pi^c(o_{i, t})$ to execute. The environment state will be transferred to $s_{t+1}$. We then can obtain $o_{i, t+1}$ and $a_{i, t+1} \thicksim \pi^c(o_{i, t+1})$ according to $s_{t+1}$.
Next, we roll back the environment simulator to the state $s_t$, and let agent $k$ reselect action $\hat{a}_{k, t} \thicksim \pi^b(o_{k,t})$ and each clean agent $i$ still use action $a_{i,t}$ to execute. The environment will enter a new state $\hat{s}_{t+1}$ and we can get $\hat{o}_{i, t+1}$, and then $\hat{a}_{i, t+1}$. Note that since the model of agents contains RNNs, we also need to roll back the hidden layer states of agents. After getting all $\hat{a}_{i, t+1}$ and $a_{i, t+1}$, we use the Equ. (\ref{equ:un}) to calculate $r^I_t$ and normalize it into the same range of values as $r^e_t$.

In summary, we hack the reward of the backdoored agent during the attack period as follows:
\begin{equation}
    r_t = (1 - \lambda)\cdot r_t^e + \lambda \cdot r_t^I
\end{equation}
where $\lambda$ represents a hyperparameter that balances the trade-off between the long-term malicious damage to the team and the short-term unilateral influence on the clean teammates.
\addtolength{\topmargin}{0.01in}


\begin{algorithm} [t!] \footnotesize
\caption{Backdoored Model Training Algorithm}
\KwIn{Network of backdoored model $\pi^b$, network of clean model $\pi^c$, an environment $Env$, clean replay buffer $\mathcal{B}_c$, poison replay buffer $\mathcal{B}_p$, spatiotemporal behavior trigger $\mathcal{T} := (\Psi, \zeta)$, attack duration $L$, poisoning rate $p$.}

\KwOut{Network of backdoored model $\pi^b$.}

Initialize $\pi^b = \pi^c$; initialize $B_c$ and $B_p$;

\For{$episode = 1$ to $max\_episodes$}{
    With probability $p$ $IsPoison = True$ and inject the backdoor trigger $\mathcal{T}$, otherwise $IsPoison = False$; $AttackDur=0$; $done = False$; initialize $\mathcal{M}$;

    \For{$t = 1$ to $episode\_limit$ and not $done$}{
    $\textbf{a}_{-k, t} = \pi^c(\textbf{o}_{-k, t})$;

    $a_{k, t} = \epsilon - greedy(\pi^b(o_{k, t}), \epsilon)$;

    $s_{t+1}, \textbf{o}_{t+1}, r_t, done = Env(a_{k, t}, \textbf{a}_{-k, t})$;

    \If{$IsPoison = True$ and $\mathcal{T}$ appears completely }{
    $AttackDur=L$;}

    \If{$AttackDur>0$}{
    Hack $r_t$ as equation (5);

    $AttackDur = AttackDur - 1$;}

    Store $(s_t, o_{k, t}, a_{k, t}, r_t, done)$ into episode memory $\mathcal{M}$;}

     \If{$IsPoison = True$}{
    Store $\mathcal{M}$ into $\mathcal{B}_p$;}

    \Else{
    Store $\mathcal{M}$ into $\mathcal{B}_c$;}

    With probability $p$ sample poison episodes from $\mathcal{B}_p$,

    otherwise sample clean episodes from $\mathcal{B}_c$;

    update $\pi^b$ with the sampled episodes;
}
\Return{$\pi^b$}.
\end{algorithm}
\setlength{\textfloatsep}{0pt}

\vspace{-2mm} 

\subsection{Backdoored Model Training}
To inject the backdoor only in a single agent, we assume that all agent models have been trained well in the clean c-MADRL environment, and during backdoor injection, we only retrain a single model and leave others to be frozen. Our complete backdoor injection procedure is outlined in Algorithm 1.

Firstly, the attacker pre-specifies a spatiotemporal behavior trigger $\mathcal{T} := (\Psi, \zeta)$ with a trigger period $\mathbb{N}^t$. Besides, an attack duration $L$ is introduced to attack only $L$ time steps after the trigger appears, which can enhance the stealthiness of the backdoor attacks. In each episode, the attacker determines whether to poison with the poisoning rate $p$. If poison, the attacker will insert the backdoor trigger (Line 3). At each time step $t$ in an episode, clean agents select their actions according to their policies, and the backdoored agent chooses a random action with the probability $\epsilon$, otherwise, it chooses an action according to its policy $\pi^b$ (Line 5-7). During the attack period $L$, the attacker will hack the reward $r_t$ as Equ (5) (Line 11). To ensure the effective training of the backdoored model, we set up two replay buffers, $\mathcal{B}_p$ and $\mathcal{B}_c$, to store poisoned and clean episodes, respectively. To update the model, we randomly sample a batch of episode data from replay buffer $\mathcal{B}_p$ or $\mathcal{B}_c$ with probability $p$ or $1-p$ (Line 14-20).

\section{Evaluation}
\subsection{Experiment Settings}
We use the StarCraft Multi-Agent Challenge (SMAC) as our experimental environment and choose 8m (8 Marines) as our test map.
All allied units (\textit{i.e.}, the red side) are controlled by c-MADRL agents, while enemy units (\textit{i.e.}, the blue side) are controlled by computer. If the enemies' health decreases, the agents gain a corresponding positive reward. When an enemy unit dies, or all enemy units die (\textit{i.e.}, win), or all allied units die (\textit{i.e.}, loss), the agents receive rewards of 10, 200 and 0 respectively.
We pick VDN and QMIX as the attacked algorithms, both of which are based on Q-value decomposition. The policy networks of clean agents trained by these two algorithms adopt the same structure with a GRU layer as follows: $|O| \rightarrow 64 \rightarrow 64 \rightleftarrows 64 \rightarrow 64 \rightarrow |A|$. Without attacks, the winning rates of VDN and QMIX based on this policy network can reach 95.6\% and 96.8\%, respectively.

We use the trigger $\mathcal{T} := (\Psi, \zeta)$ defined in Example 1 to evaluate our backdoor attacks. To inject the backdoor, we set the size of both the clean replay buffer $\mathcal{B}_{c}$ and the poison replay buffer $\mathcal{B}_{p}$ to 5000, the size of batch to 32, the discount factor $\gamma=0.99$, the greedy factor $\epsilon=0.05$, the learning rate $\alpha=5e-4$, and the update frequency of the target network to 200 steps. In each poisoned episode, we search if there is an observable enemy unit whose position relative to the allied unit controlled by the backdoored agent satisfies the first spatial constraint in $\Psi$. If found, we will take control of the enemy unit to behave following $\zeta$ defined in $\mathcal{T}$, and return it to its heuristic controller after the trigger. Additionally, we set the poisoning rate $p=0.05$, the attack period $L = 20$ and $\lambda=0.5$ in the hacked reward.

\subsection{Numerous Results}

\begin{figure}[t]
    \captionsetup{font=small}
    \centering
    \begin{subfigure}{.245\textwidth}
        \centering
        \includegraphics[width=\linewidth]{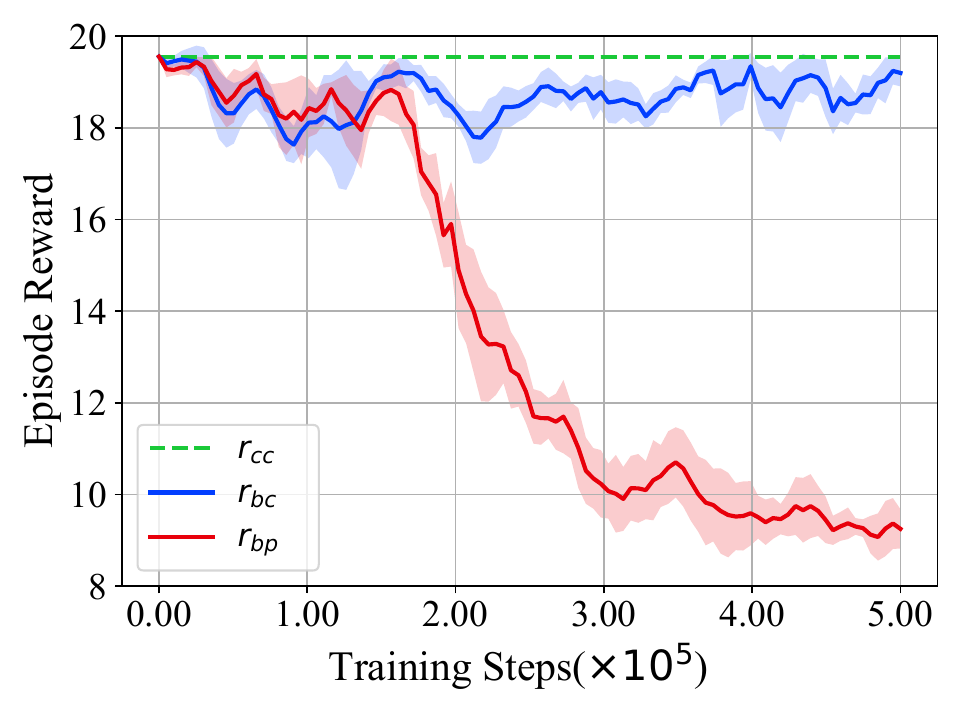}
        \caption{Episode reward (VDN)}

    \end{subfigure}%
    \begin{subfigure}{.245\textwidth}
        \centering
        \includegraphics[width=\linewidth]{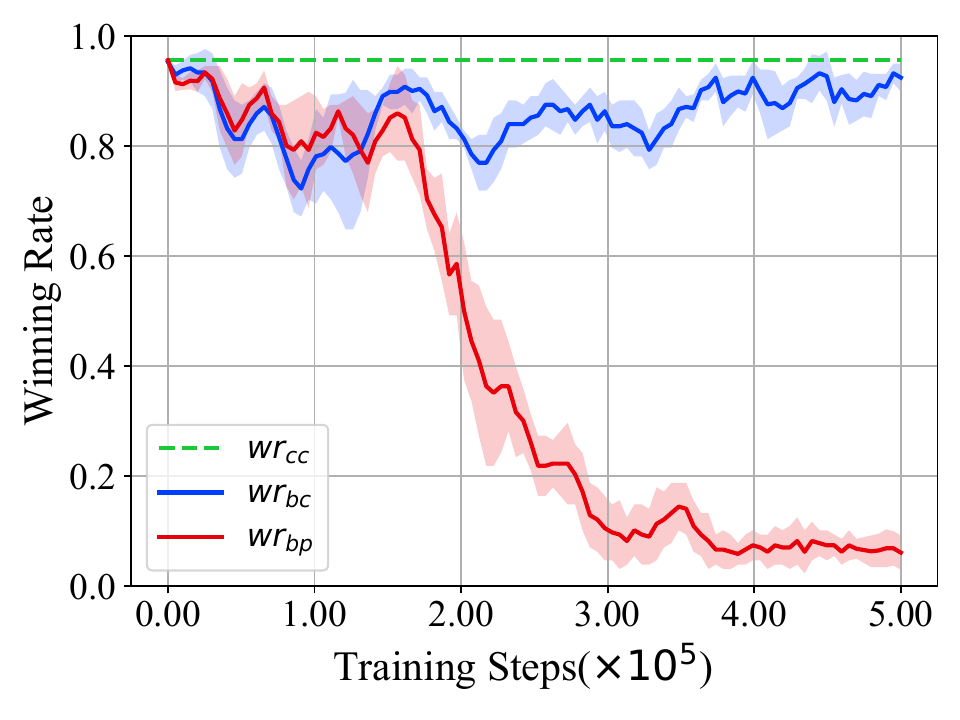}
        \caption{Winning rate (VDN)}

    \end{subfigure}

    \medskip 

    \begin{subfigure}{.245\textwidth}
        \centering
        \includegraphics[width=\linewidth]{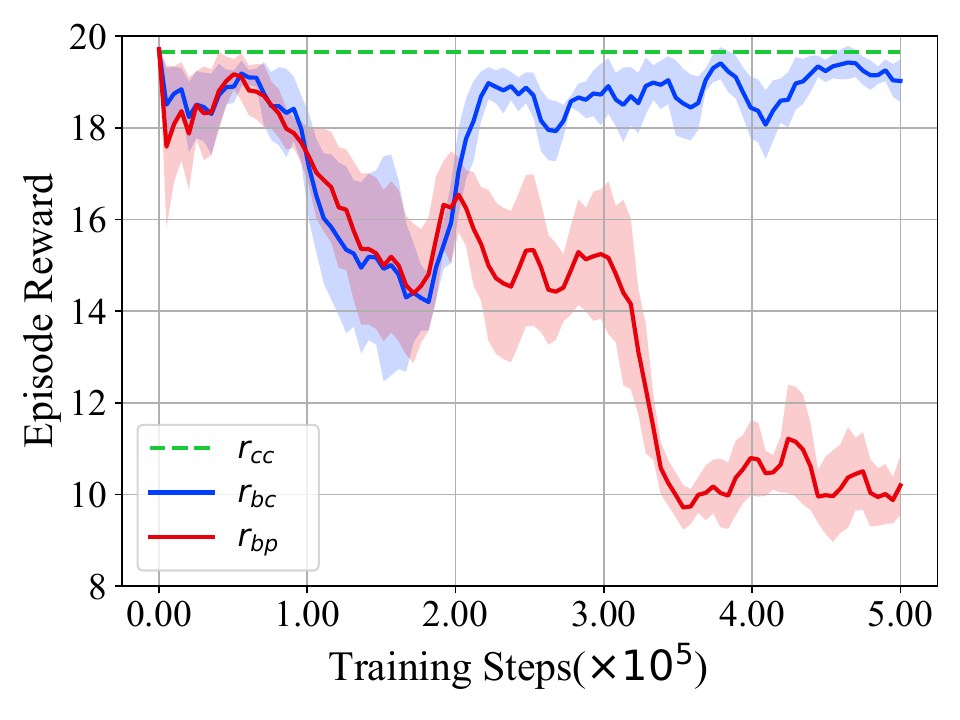}
        \caption{Episode reward (QMIX)}
    \end{subfigure}%
    \begin{subfigure}{.245\textwidth}
        \centering
        \includegraphics[width=\linewidth]{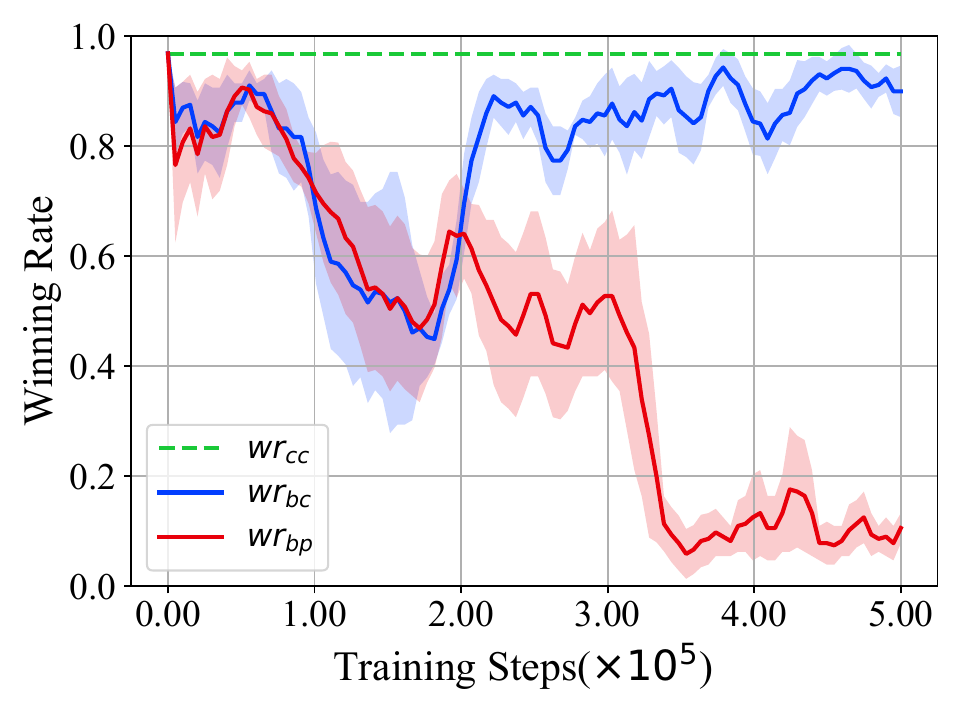}
        \caption{Winning rate (QMIX)}
    \end{subfigure}
    \caption{The episode rewards and winning rates of the backdoored models attacking VDN and QMIX. }
    \label{fig:images}
    \vspace{2mm}
\end{figure}

Fig. 2 illustrates the training process of the backdoored models attacking VDN and QMIX. We first use the team's average episode reward and winning rate in clean episodes (without the trigger $\mathcal{T}$) and poisoned episodes (with the trigger $\mathcal{T}$) to evaluate the performance of the backdoored models, respectively. In Fig. 2, $r_{bc}$, $r_{bp}$, and $wr_{bc}$, $wr_{bp}$ represent the episode rewards and winning rates of the backdoored model in clean and poisoned episodes, respectively, and additionally $r_{cc}$ and $wr_{cc}$ represent the episode reward and winning rate of the original clean model in clean episodes. We can find that both for VDN and QMIX, $r_{bp}$ and $wr_{bp}$ gradually decrease with training and finally converge to below 10.0 and below 0.10. Besides, $r_{bc}$ ($wr_{bc}$) decreases slightly at first, but picks up later and finally converges to above 19.0 (0.90), which is slightly lower than $r_{cc}$ ($wr_{cc}$). This result shows that our backdoor attacks are effective both against VDN and QMIX, \textit{i.e.}, the malicious actions taken by the backdoored agent after the trigger is present can cause the entire team to fail.

\begin{figure}[t]
    \vspace{1mm}
    \captionsetup{font=small}
    \centering
    \begin{subfigure}{.24\textwidth}
        \centering
        \includegraphics[width=\linewidth]{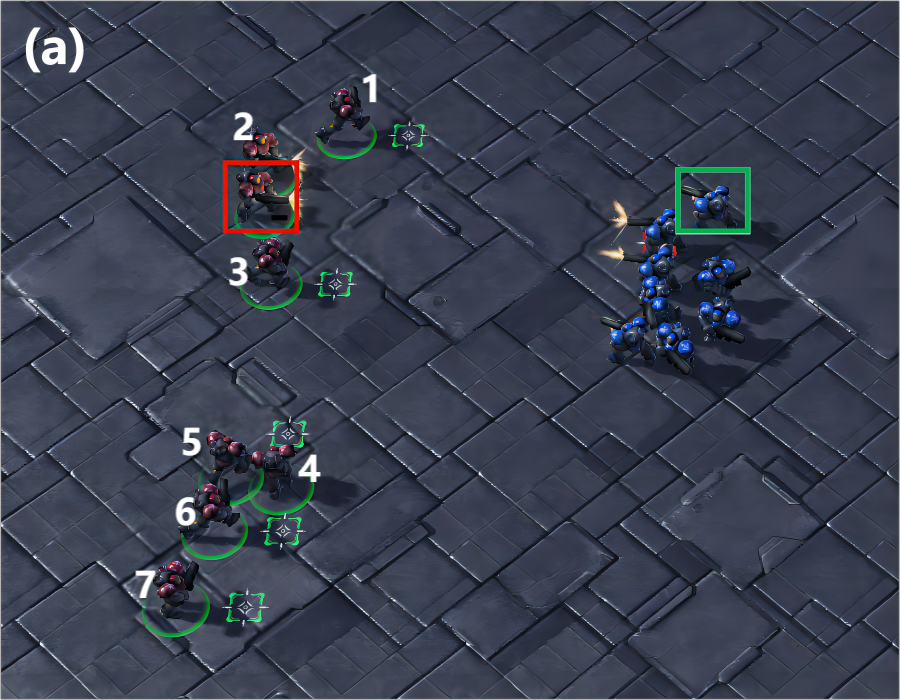}
    \end{subfigure}%
    \hfill 
    \begin{subfigure}{.24\textwidth}
        \centering
        \includegraphics[width=\linewidth]{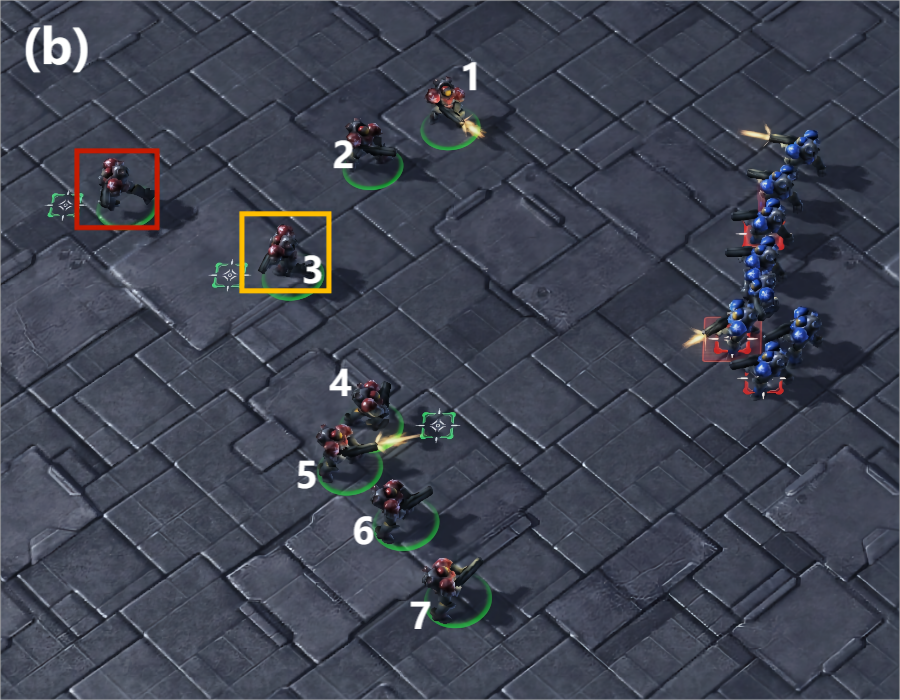}
    \end{subfigure}

    \medskip 

    \begin{subfigure}{.24\textwidth}
        \centering
        \includegraphics[width=\linewidth]{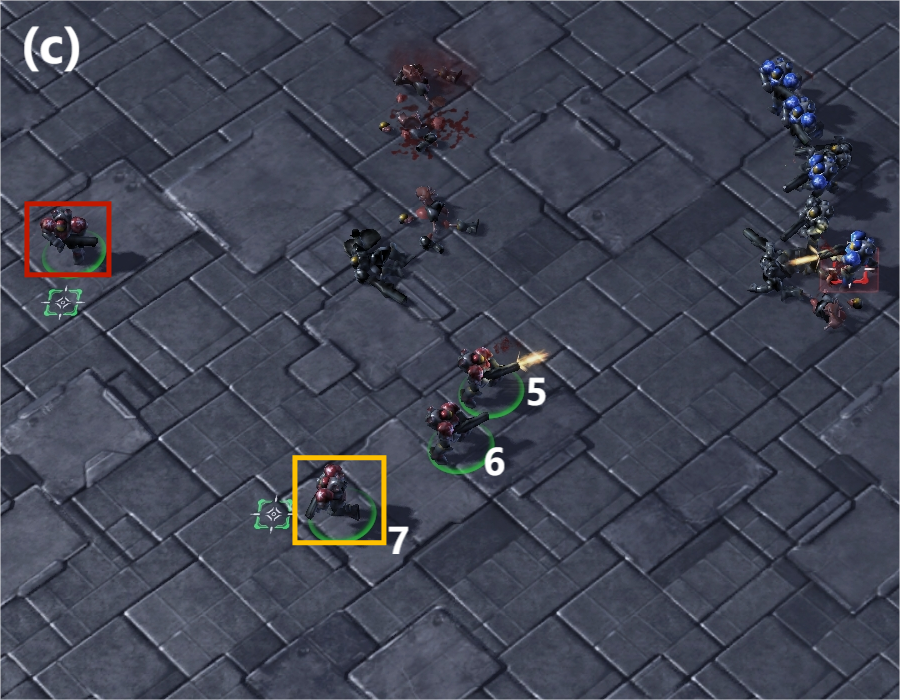}
    \end{subfigure}%
    \hfill 
    \hspace{.0001\textwidth}
    \begin{subfigure}{.24\textwidth}
        \centering
        \includegraphics[width=\linewidth]{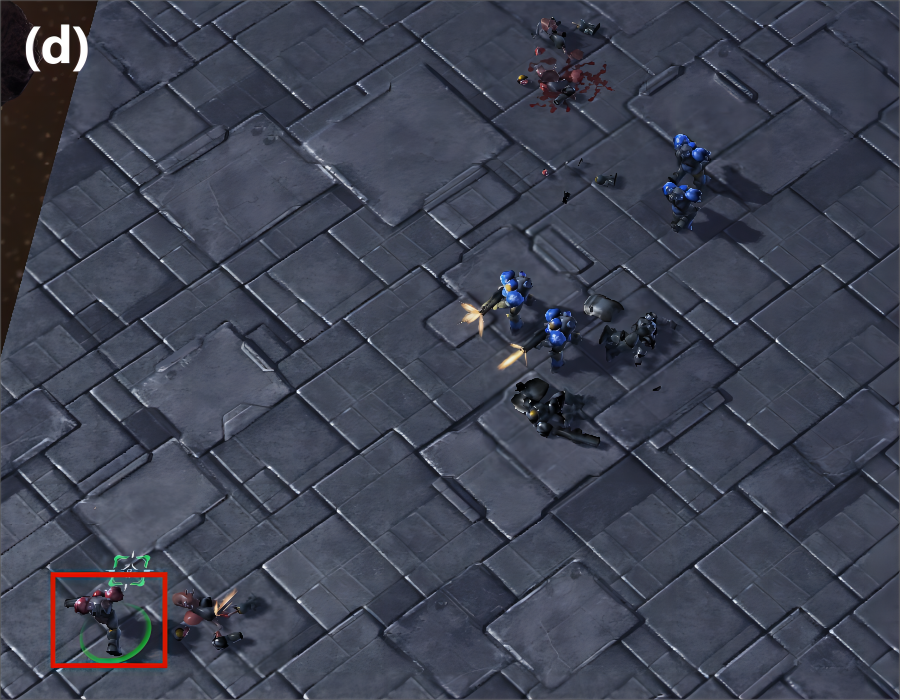}
    \end{subfigure}
    \caption{The behaviors of all agents in an attack period. The trigger is shown in the green rectangle, the backdoored agent is shown in the red rectangle and affected clean agents are shown in yellow rectangles. Numbers represent the ID of clean agents.}
    \vspace{-2mm}
\end{figure}

\begin{figure}[t]
    \captionsetup{font=small}
    \centering
    \begin{subfigure}{.24\textwidth}
        \centering
        \includegraphics[width=\linewidth]{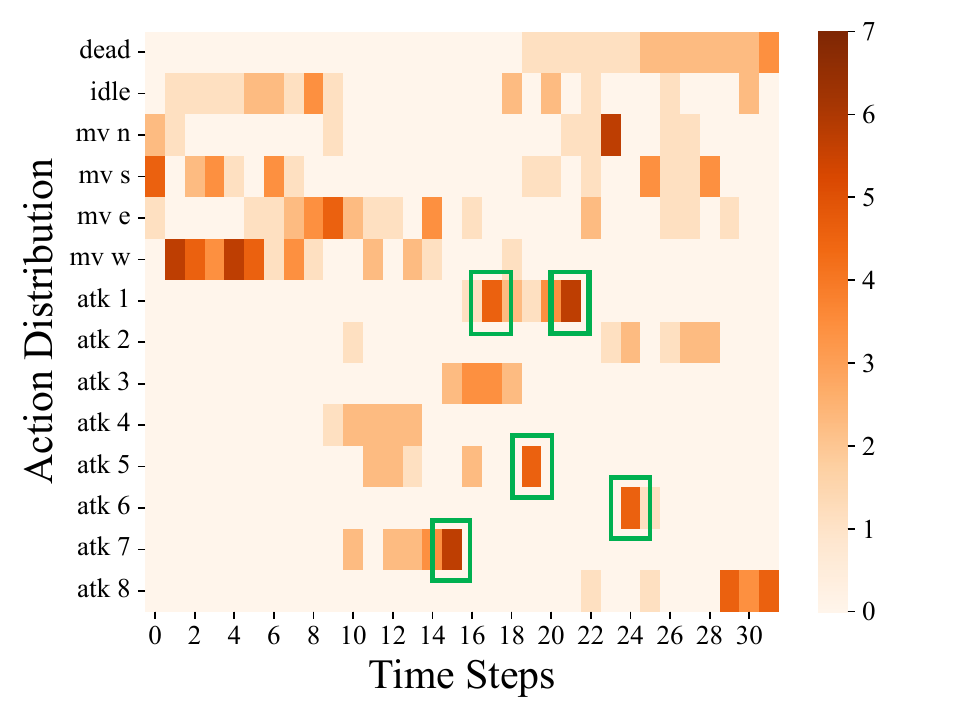}
        \caption{Clean episode}
    \end{subfigure}%
    \begin{subfigure}{.24\textwidth}
        \centering
        \includegraphics[width=\linewidth]{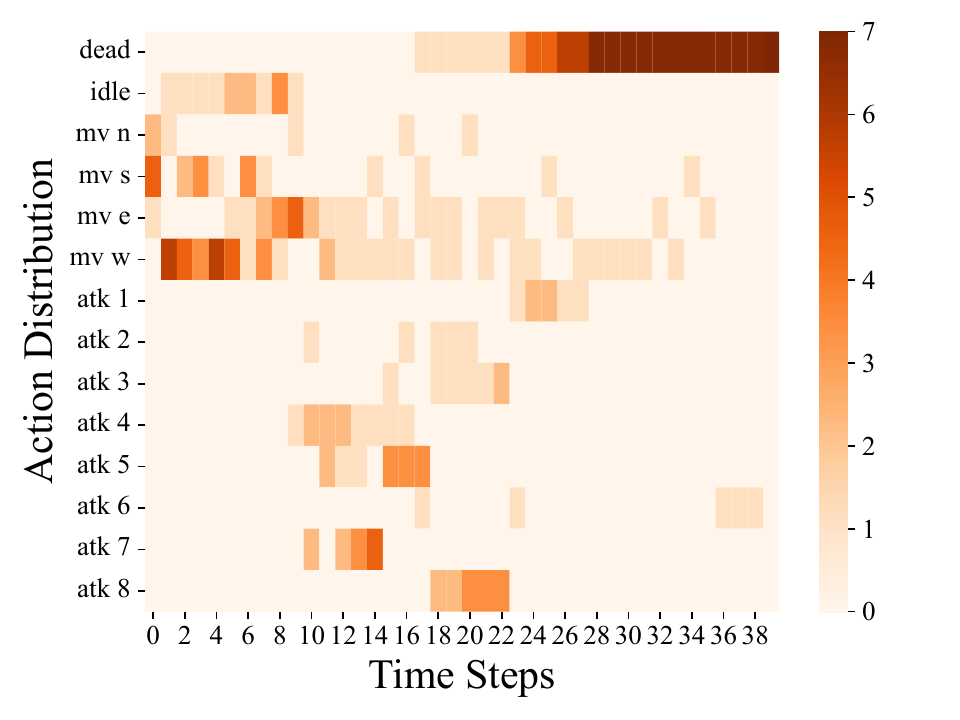}
        \caption{Poisoned episode}
    \end{subfigure}
    \caption{Action distribution of all clean agents in a clean episode and a poisoned episode. The green rectangle represents the behavior of focused fire.}
    \vspace{2mm} 
\end{figure}

To explore how the backdoored agent can cause the team to fail, we analyze the behaviors of the backdoored agent during the attack period and show the results in Fig. 3. Before the attack begins, the agents take scattered positions to attack the enemy, which is the strategy learned through QMIX, as shown in Fig. 3 (a). We control an enemy unit around the backdoored allied unit to perform trigger actions.
Once the trigger occurs completely, as shown in Fig. 3 (a), the backdoored allied unit starts to move away from the main battle but moves within the range of its teammates' vision, successfully inducing some of its teammates (3 and 7) to move to a disadvantageous position or change from attacking behaviors to moving behaviors, causing the remaining teammates to encounter enemy attacks, as shown in Fig. 3 (b-c). Ultimately, all other teammates are killed, followed by the backdoored agent itself, resulting in a game loss, as shown in Fig. 3 (d). Fig. 4 further illustrates the difference in the joint action distribution for all clean agents in a clean and a poisoned episode, respectively. We can find that in the clean episode, the agents perform the behavior of focused fire, rapidly killing enemy units one by one, and thus achieving victory. At the end (time step 31), there are still four agents alive. In the poisoned episode, before the attack begins, the agents have the same action distribution as in the clean episode. During the attack period (between time steps 11 and 30), the agents are influenced by the backdoored agent and begin to exhibit deviant behaviors, including no longer performing focused fire, a more spread out distribution of actions, a decrease in attacking actions, and an increase in moving actions. These changes lead to the death of all agents and the failure of the allied team.


\begin{table}[t]
    \captionsetup{font=small}
    \centering
    \vspace{4mm}
    \caption{Attack performance with different $\lambda$.}
    \label{tab:perf}
    \renewcommand{\arraystretch}{1.25}
    \setlength{\tabcolsep}{5pt}
    \begin{tabular}{ccccccc}
        \toprule 
        Alg & metric & $\lambda=0$ & $\lambda=0.2$ & $\lambda=0.5$ & $\lambda=0.8$ & $\lambda=1$ \\
        \hline
        \multirow{2}{*}{VDN} & $CPVR$ & 3.8\% & 4.8\% & 3.7\% & 2.7\% & 1.7\%\\
         & $ASR$ & 79.1\% & 90.6\% & 91.6\% & 68.6\% & 13.2\%\\
        \hline
        \multirow{2}{*}{QMIX} & $CPVR$ & 3.9\% & 5.9\% & 3.9\% & 6.1\% & 1.8\%\\
         & $ASR$ & 69.0\% & 82.4\% & 90.7\% & 70.0\% & 11.2\%\\
        \bottomrule 
    \end{tabular}
\end{table}

We further evaluate the performance of our backdoor attacks with different values of $\lambda \in \{0, 0.2, 0.5, 0.8, 1\}$ in hacked reward function, as shown in Table \ref{tab:perf}. We use two metrics to evaluate the attack performance: clean performance variance rate ($CPVR = |wr_{bc} - wr_{cc}| / wr_{cc}$) and attack success rate ($ASR = |wr_{bp} - wr_{cc}| / wr_{cc}$).
A smaller $CPVR$ indicates that the backdoor is better hidden and a larger $ASR$ indicates that the attack is more effective.
We can find that when $\lambda = 0$, the $ASR$ on VDN and QMIX can reach up to 79.1\% and 69.0\%, respectively. This indicates that including only the first term in the hacked reward, \textit{i.e.}, reversing the original reward is useful to some extent against c-MADRL. However, when $\lambda = 1$, the $ASR$ is only 13.2\% on VDN and 11.2\% on QMIX, although it can maintain a very low $CPVR$. This suggests that the attack is not very effective when considering only the second term in the hacked reward, i.e., the effect of the backdoored agent on the next time-step actions of its teammates. Attacks perform best when $\lambda = 0.5$, being able to maintain a 3.7\% $CPVR$ and achieve a 91.6\% $ASR$ on VDN, and maintain a 3.9\% $CPVR$ and achieve a 90.7\% $ASR$ on QMIX. This further demonstrates the effectiveness of our hacked reward for backdoor attacks against c-MADRL.

\section{Conclusions And Future}
In this paper, we study backdoors in c-MADRL. To enhance the stealthiness, effectiveness, and practicality of backdoor attacks, we propose a novel backdoor attack against c-MADRL, which can disrupt the entire multi-agent team by implanting a backdoor in only one agent. We use spatiotemporal features rather than an instant state as the backdoor trigger, and design the malicious reward function according to the characteristics of c-MADRL. We evaluate the proposed backdoor attacks on VDN and QMIX algorithms in SMAC, and the experimental results demonstrate that the backdoor attacks can achieve great attack success rate and low clean performance variance rate. In the future, we will explore backdoor attacks in black-box scenarios, as well as study effective defence methods for c-MADRL backdoors.
\vspace{-0.1cm}

\section*{Acknowledgement}
This work was supported in part by the National Natural Science Foundation of China 62202387, in part by the GuangDong Basic and Applied Basic Research Foundation under Grant 2021A1515110279, and in part by the Key Research and Development Program of Shaanxi under Grant 2022GXLH-02-03.
\vspace{-0.2cm}

\bibliographystyle{IEEEtran}
\bibliography{IEEEabrv, ref}
\end{document}